%% file: vita_slam_cbs_19.tex
\documentclass[letterpaper, 10 pt, fleqn, conference, twoside]{ieeeconf}  
\IEEEoverridecommandlockouts                              
\usepackage{capt-of,graphicx,subcaption}
\usepackage{xargs}                      
\usepackage[pdftex,dvipsnames,table,xcdraw]{xcolor}  
\usepackage{multirow}
\usepackage[colorinlistoftodos,prependcaption]{todonotes}
\usepackage[hidelinks]{hyperref} 
\hypersetup{
  colorlinks   = true, 
  urlcolor     = blue, 
  linkcolor    = blue, 
  citecolor   = red 
}

\usepackage{amsmath,amssymb,bm,mathtools}
\DeclareMathOperator*{\argmin}{arg\,min}

\usepackage{cite}
\usepackage{multirow}
\usepackage[font={small,it}]{caption}
\usepackage{relsize}
\usepackage{tikz} 
\usetikzlibrary{shadings}
\usepackage[margin=1.5cm]{geometry} 
\usepackage{algorithm}
\usepackage[noend]{algpseudocode}
\usepackage[subpreambles]{standalone} 

\usepackage{array}
\usepackage{booktabs}
\usepackage[utf8]{inputenc}
\usepackage{pgfplots}
\pgfplotsset{compat=newest}
\usepgfplotslibrary{groupplots}
\usepgfplotslibrary{dateplot}

\makeatletter
\def\BState{\State\hskip-\ALG@thistlm}
\makeatother

\newcommandx{\os}[2][1=]{\todo[linecolor=red,backgroundcolor=red!25,bordercolor=red,#1,inline]{OS: #2}} 
\newcommandx{\mt}[2][1=]{\todo[linecolor=Orchid,backgroundcolor=Goldenrod!25,bordercolor=Orchid,#1,inline]{MP: #2}} 
\newcommandx{\vk}[2][1=]{\todo[linecolor=blue,backgroundcolor=blue!25,bordercolor=blue,#1,inline]{VK: #2}} 
\newcommandx{\kt}[2][1=]{\todo[linecolor=OliveGreen,backgroundcolor=OliveGreen!25,bordercolor=OliveGreen,#1,inline]{KT: #2}} 

\graphicspath{{./sections/figures/}} 

\title{\LARGE \textbf{ViTa-SLAM: A Bio-inspired Visuo-Tactile SLAM for Navigation while Interacting with Aliased Environments}}

\author{Oliver Struckmeier$^{*}$, Kshitij Tiwari$^{*}$, Mohammed Salman, Martin J. Pearson, and Ville Kyrki
\thanks{$^{*}$ The authors have equal contribution.}
\thanks{This research has received funding from the European Union’s Horizon 2020 Framework Programme for Research and Innovation under the Specific Grant Agreement No. 785907 (Human Brain Project SGA2).}
\thanks{Oliver Struckmeier, Kshitij Tiwari, and Ville Kyrki are with the Department of Electrical Engineering and Automation, Aalto University, Espoo 02150, Finland \tt \footnotesize \{firstname.lastname\}@aalto.fi}
\thanks{Mohammed Salman is with the Bristol Robotics Laboratory, University of Bristol and University of the West of England, Bristol, UK (\tt\footnotesize ms13417@bristol.ac.uk)}
\thanks{Martin J. Pearson is with the Bristol Robotics Laboratory, Bristol BS16 1QY, U.K (\tt\footnotesize martin.pearson@brl.ac.uk)}
}

\begin{document}
\maketitle
\input{sections/abstract}

\input{sections/intro}

\input{sections/methods}

\input{sections/experiments}

\input{sections/performance}

\input{sections/conclusion}

\bibliographystyle{ieeetr}
\bibliography{vita_slam_cbs_19}%
\nocite{rusu2008learning,kim2007biomimetic}
\end{document}

%% file: sections/abstract.tex
\begin{abstract}
RatSLAM is a rat hippocampus-inspired visual Simultaneous Localization and Mapping (SLAM) framework capable of generating semi-metric topological representations of indoor and outdoor environments. Whisker-RatSLAM is a $6$D extension of the RatSLAM and primarily focuses on object recognition by generating point clouds of objects based on tactile information from an array of biomimetic whiskers. This paper introduces a novel extension to both former works that is referred to as \textit{ViTa-SLAM} that harnesses both vision and tactile information for performing SLAM. This not only allows the robot to perform natural interactions with the environment whilst navigating, as is normally seen in nature, but also provides a mechanism to fuse non-unique tactile and unique visual data. Compared to the former works, our approach can handle ambiguous scenes in which one sensor alone is not capable of identifying false-positive loop-closures.
\end{abstract}

%% file: sections/intro.tex
\section{Introduction}
\begin{figure}[!htb]
\centering
\begin{subfigure}[t]{.5\textwidth}
  \centering
        \includegraphics[scale=0.25]{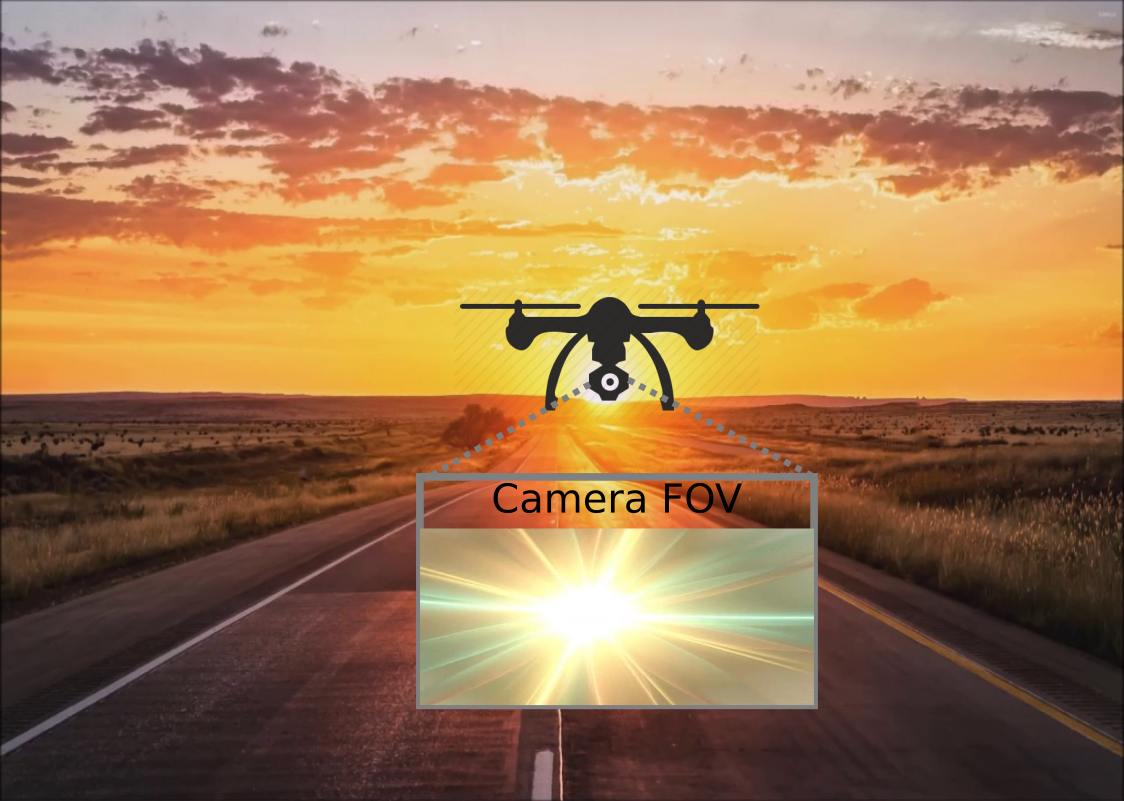}
        \caption{Sudden lighting changes.}
        \label{fig:robotic_motivation}
\end{subfigure}%
\linebreak
\begin{subfigure}[t]{.5\textwidth}
  \centering
  \includegraphics[scale=0.25]{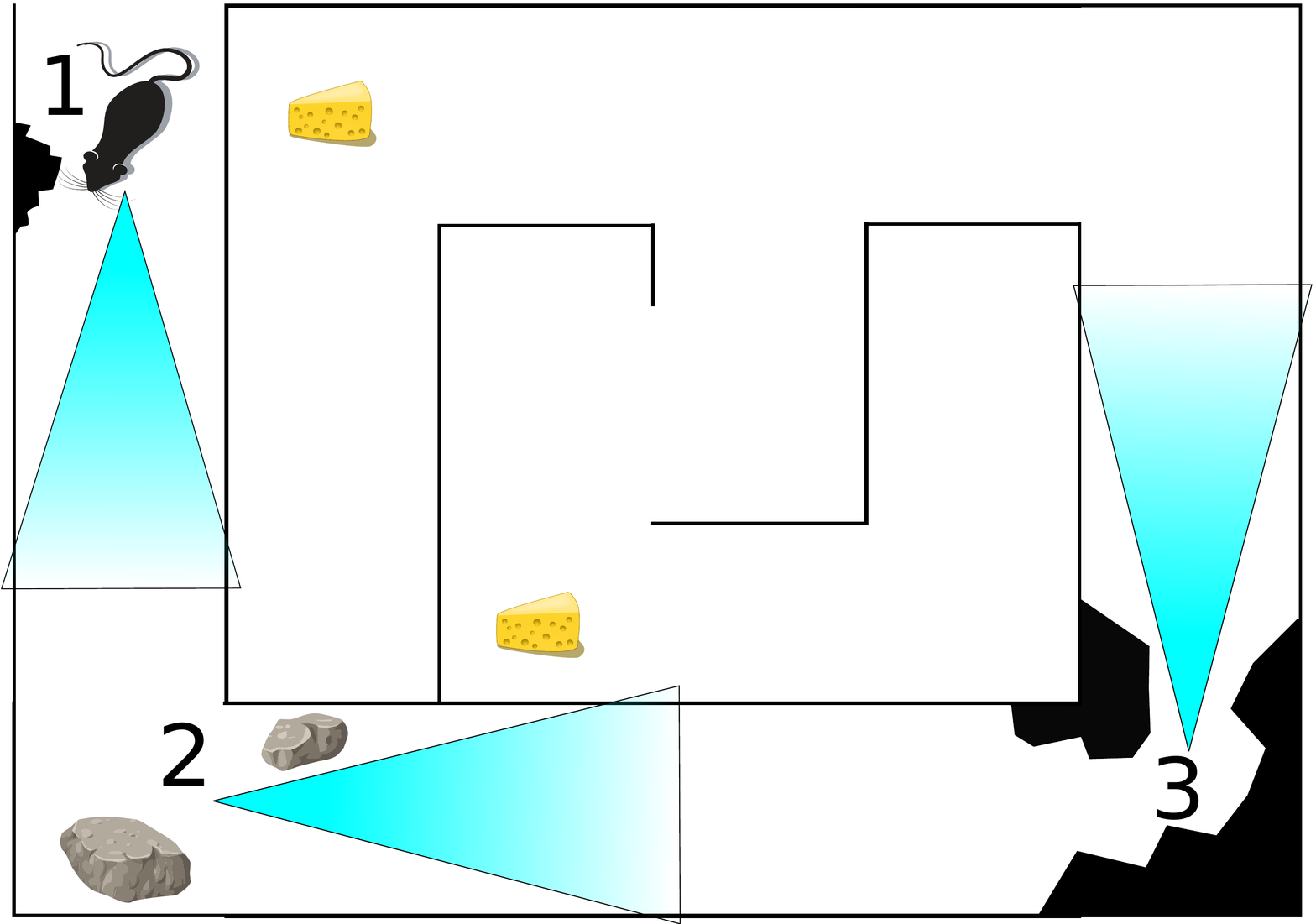}
  \caption{Rat in visually ambiguous maze.}
  \label{fig:biological_motivation}
\end{subfigure}
\caption{Multiple sensory modalities can help in situations in which one sensor alone is not sufficient. For example, a sudden flash of sunlight can blind a drone's camera in flight making visual navigation impossible. A rat navigating through a maze sees the same visual scene in multiple locations (marked as $1,2,3$). Corridors and corners in the maze can look the same, especially considering the poor acuity of rodent vision. The blue polygons represent indicative fields of view to highlight this ambiguity. Tactile sensing can help the rat to distinguish ambiguous scenes in these situations.}
\label{fig:motivation}
\vspace{-0.5cm}
\end{figure}

Robots are often equipped with off-the-shelf sensors like cameras which are used for navigation, however, vision is sensitive to extremes in lighting conditions such as shadows or unpredictable changes in intensity as shown in Fig.~\ref{fig:robotic_motivation}. Whilst other on-board sensors like laser range finders can be used in such situations they too are impaired by reflective and absorbing surfaces. Similarly, sensory systems as they occur in nature are subject to impairments, \textit{e.g.,} a rat moving through a maze in ill-lit conditions as illustrated in Fig.~\ref{fig:biological_motivation}. However, through the process of evolution nature has equipped animals to gracefully accommodate such scenarios. Given the coarse vision and challenges of a rodent's natural environment, they are known to rely on tactile feedback derived form whiskers aside from vision to decipher their own location. Considering the example depicted in Fig.~\ref{fig:biological_motivation}, a rat navigates a maze where in certain locations visual or tactile information is ambiguous but combining tactile and visual information can help to discern similar locations. Conventional robots lack such a robust capability to interact with their environment through contact. Thus, biomimetic robots are gaining traction \cite{prescott2009whisking} which has now made it possible to harness visual and tactile sensory modalities for informed decision making. However, it still remains unclear how to best process and combine information from disparate sensory modalities to aid in spatial navigation.

Previous works on visuo-tactile sensor fusion like~\cite{alt2014navigation,alt2013haptic} usually combine sensory modalities of varying sensing ranges. The key requirement of these methods was the need to have a redundant field-of-view. 
Other works in this domain like ~\cite{bhattacharjee2015combining,pinto2016curious,DBLP:journals/ral/CalandraOJLYMAL18} have mainly focused on creating dense haptic object/scene maps. Whilst these methods allow for environmental interactions, they are primarily designed for tactile object exploration and grasping. Although, tactile sensing is increasingly being used in these domains, it remains yet to be applied for performing Simultaneous Localization and Mapping (SLAM).

In the context of SLAM, previous works have demonstrated the strengths of a bio-inspired SLAM system and shown its application using single-sensory modalities such as either vision~\cite{ball2013openratslam}, sonar~\cite{steckel2013batslam} or WiFi~\cite{berkvens2014biologically}. However, such methods rely on the uniqueness of the data and are thus susceptible to false-positive place recognition. This problem was previously addressed by fusing information from an array of active sensors each providing rich information~\cite{milford2013brain}. Despite the robustness to illumination changes, this method is not capable of fusing non-unique sensory information.

To address the challenges of place recognition in aliased environments using multiple sensors we present our novel method of identifying and preventing false-positive place recognition by combining long-range (unique) vision, with short-range (non-unique) tactile information. Additionally, the proposed method does not rely on sensory redundancy. Our preliminary results presented in \cite{struckmeier2019vita} showed that the method presented herewith is capable of preventing false-positive place recognition from a vision-only SLAM system. Subsequently, a robust sensor fusion algorithm has been developed to integrate information from unique and non-unique sensory modalities such as cameras and whiskers, respectively. Additionally, performance metrics are presented herewith to compare and evaluate model performance against vision or tactile only sensing.

%% file: sections/methods.tex
\section{Bio-inspired SLAM}
This work draws inspiration from two well-known bio-inspired SLAM frameworks: \textit{RatSLAM,} a rat hippocampal model based visual SLAM architecture \cite{ball2013openratslam}; and \textit{Whisker-RatSLAM,} an extension of RatSLAM aimed primarily at tactile object exploration \cite{salman2018whisker}. This work relies on modified variants of these models referred to as \textit{Visual-SLAM} and \textit{Tactile-SLAM}. In this section these models are summarized and the differences from the works in \cite{ball2013openratslam,salman2018whisker} are shown. Lastly, the proposed \textit{ViTa-SLAM} model is introduced.

For each of the models, an overall system architecture is provided using the following convention: nodes represented by right isosceles triangles represent raw sensory data; nodes represented by ellipse(s) represent pre-processing of sensory data before they are converted to input features represented by rounded boxes. The outputs from the models are represented by regular boxes, the pre-processing and feature generation stages are highlighted in light blue and light red blocks, respectively.

\subsection{Visual-SLAM}
\begin{figure}[!htbp]
\centering
\resizebox{.5\textwidth}{!}{\input{sections/figures/vanilla_ratslam}}
\caption{The overview over the visual-SLAM implementation used in this work.}
\label{fig:OpenRatSLAM_Strucutre}
\end{figure}
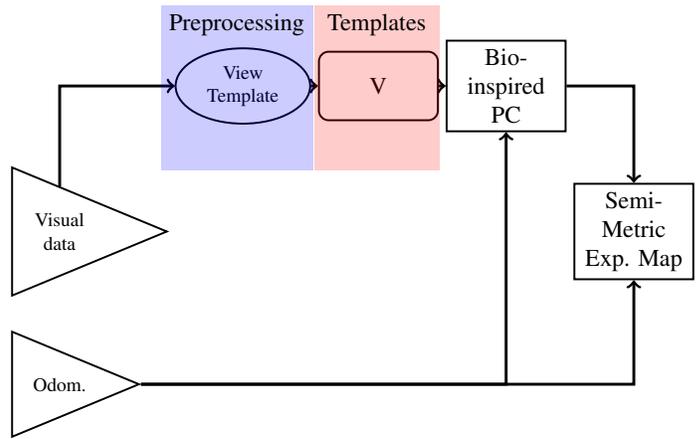

When investigating the way rodents navigate from a bio-inspired perspective, RatSLAM as introduced in \cite{ball2013openratslam,milford2008robot,milford2008mapping,milford2013brain,quigley2009ros}, has been proven to be a capable visual SLAM method. RatSLAM is loosely based on the neural processes underlying navigation in the rodent (primarily rat) brain, more specifically the hippocampus. Fig.~\ref{fig:OpenRatSLAM_Strucutre} shows an overview of the visual-SLAM implementation used in this work.

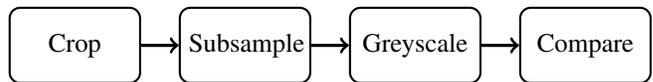
\begin{figure}[!htbp]
\centering
\input{sections/figures/lv_flow}
\caption{View template pre-processing}
\label{fig:view_template}
\end{figure}

During the \textit{preprocessing} phase, the input of a camera (visual data) is downsampled to reduce computational cost and to simulate the coarse vision of rats. In this process the incoming visual data is cropped to remove areas that do not provide unique features, like for example the ground. The cropped image is subsampled and converted to greyscale as shown in Fig.~\ref{fig:view_template}. 

The preprocessed sensory information is now parsed through $3$ major components of the RatSLAM architecture:
\begin{itemize}
\item Pose Cells
\item Local View Cells
\item Experience Map \newline
\end{itemize}

The \textit{pose cells} \cite{ball2013openratslam} encode the robot's current best pose estimate. Pose cells are represented by a \textit{Continuous Attractor Network} (CAN) \cite[Ch.~4]{milford2008robot}, the posecell network, to resemble the \textit{grid cells} as introduced in \cite{hafting2005microstructure}. The grid cells are neurons found in many mammals and are shown to be used in navigation. In the $3$D posecell network, the robot's pose estimate ($x, y$ position and heading angle $\gamma$) is encoded as a energy packet that is moved through energy injection based on odometry and place recognition.

The \textit{local view (LV) cells} are an expandable array of units used to store the distinct visual scenes as a \textit{visual template} in the environment using a low resolution subsampled frame of pixels. The visual template generated from the current view is compared to all existing view templates by shifting them relative to each other. If the current view is novel, a local view cell is linked with the centroid of the dominant activity package in the pose cells at the time when a scene is observed. When a scene is seen again, the local view cell injects activity into the pose cells.

The \textit{experience map} is a semi-metric topological representation of the robot's path in the environment generated by combining information from the pose cells and local view cells into \textit{experiences}. Each experience is related to the pose cell and local view cell networks via the following 4-tuple: $<x,y,\gamma,V>$ where $x,y,\gamma$ represent the location of the cell in the PC network while $V$ corresponds to the view associated with the LV cell that relates to the queried experience \cite{milford2005experience}.

Initially the robot relies on odometry which is subject to an accumulating error. When loop closure events happen, meaning a scene has been seen already, the pose estimate based on the odometry is compared to the pose of the experience and graph relaxation is applied~\cite{milford2005experience}.

The following differences to RatSLAM have been introduced in the visual-SLAM implementation: first, we use odometry from the robot instead of visual odometry as was originally done to determine the translational and rotational speeds of the robot. Second, the method of template matching and generation has been modified to account for multiple sensory modalities. Third, the posecell network (PC) is now capable of handling a wider range of robot motion such as moving sideways.

\subsection{Tactile-SLAM}
Whisker-RatSLAM is a $6$D tactile SLAM algorithm inspired by RatSLAM. Instead of taking input from a camera, it uses a tactile whisker-array mounted on a robot as its only sensor \cite{Pipe2016,pearson2013simultaneous}. The whisker-array consists of $6 \times 4$ whiskers, each capable of measuring the point of whisker contact in $3$D space, and the $2$D deflection force at their base \cite{sullivan2012tactile}. Whisker-RatSLAM \cite{salman2018whisker} has been demonstrated for mapping objects and localizing the whisker array relative to the surface of an object. Similar to RatSLAM, Whisker-RatSLAM generates a semi-metric topological map, the \textit{object exploration map}, which contains complex $6$DOF experience nodes.

\begin{figure}[!htbp]
\centering
\resizebox{.5\textwidth}{!}{\input{sections/figures/6dof_ratslam}}
\caption{The overview over the tactile-SLAM implementation used in this work.}
\label{fig:6dof_ratslam}
\end{figure}
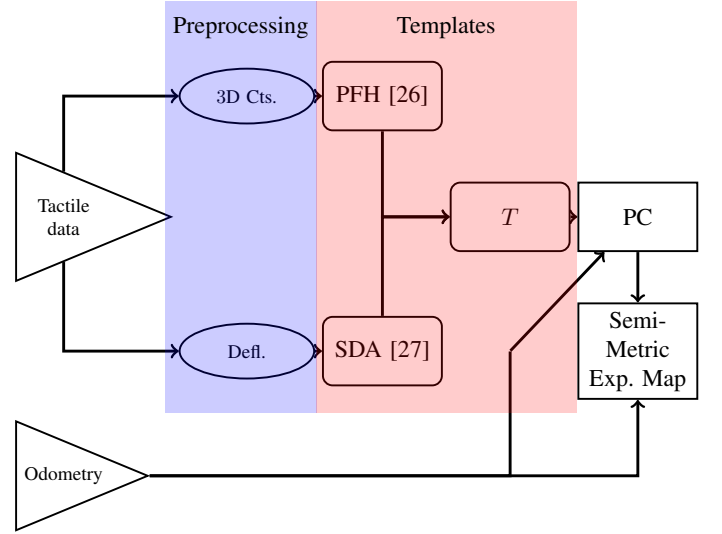

In \cite{salman2018whisker}, the authors proposed combining these object exploration maps with simple $3$DOF experience map generated using RatSLAM with the whisker-input resulting in a topological terrain exploration map with two different types of experience nodes. Fig. \ref{fig:6dof_ratslam} shows an overview of the Whisker-RatSLAM algorithm. The tactile data acquired by whisking encompasses $3$D contact point cloud of the object ($3$D Cts.) and the deflection data (Defl.). The point cloud is used to generate the \textit{Point Feature Histogram (PFH)} while the deflection data is used to generate \textit{Slope Distribution Array (SDA)}. Both PFH and SDA are then fused to obtain a $6$D Feature Cell (FC). Similar to the RatSLAM experience map, the pose grid cells and FC that were active at a specific 6D pose of the whisker-array are associated with each other and combined into \textit{experience nodes}.  The experience in this case is defined as the 7-tuple: $<x,y,z,\alpha,\beta,\gamma,F>$ where $x,y,z,\alpha,\beta,\gamma$ represents the 6D pose including euler angles for orientation and $F \gets \{PFH \cup SDA \}$ represents the features associated with that experience. The experience node form the \textit{object exploration map (Obj. Expl. Map)}. In order to adapt the activation of the pose cell in accord with the robot motion, the odometry information is also used in the pose grid.

The tactile-SLAM implementation is based on Whisker-RatSLAM, but instead of a $6$D posecell network this work uses the same $3$D posecell network as the visual-SLAM implementation to allow compatibility and to reduce computation cost for navigation in $3$D space. Furthermore, the tactile-SLAM implementation used in this work does not use feature cells, but instead combines the SDA and PFH data into $3$D tactile template that are used in a similar way as $3$D visual templates. Fig. \ref{fig:6dof_ratslam} shows an overview of the tactile-SLAM algorithm. The tactile data acquired by whisking encompasses $3$D contact point cloud of the object ($3$D Cts.) and the deflection data (Defl.). The point cloud is used to generate the \textit{Point Feature Histogram (PFH)} while the deflection data is used to generate \textit{Slope Distribution Array (SDA)}. Both PFH and SDA are then fused to obtain a tactile template (T). Similar to the RatSLAM experience map, the pose grid cells and T that were active at a specific pose of the whisker-array, are associated with each other and combined into \textit{experience nodes}. The experiences are, opposed to the 7-tuple used in Whisker-RatSLAM, defined as a 4-tuple: $<x,y,\gamma,T>$ and $T \gets \{PFH \cup SDA \}$ represents the tactile template associated with that experience. The experience nodes also form a semi-metric experience map similar to the visual-SLAM method. Similar to the visual-SLAM method, the robot's odometry information is also used to move the pose grid.

To generate tactile information using whiskers, one challenge is how to control the whisker-array in order to improve the quality of the sensory information. Previous research on rats \cite{grant2009active} has identified a number of whisking strategies that rodents use to potentially improve the sensory information they obtain. One of these strategies is called \textit{Rapid Cessation of Protraction} (RCP) and refers to the rapid reduction in motor drive applied to the whisker when it makes contact with an object during the protraction phase of exploratory whisking \cite{mitchinson2007feedback}. This effectively reduces the magnitude of bend of the whisker upon contact which in artificial arrays, such as shown in \cite{salman2016advancing}, improves the quality of sensory information by constraining the range of sensory response to a region best suited for signal processing. Furthermore, damage to the whiskers from contact is significantly reduced. 

As opposed to full $6$D pose estimation in Whisker-RatSLAM, the modified tactile-SLAM estimates only the $3$D pose ($x,y,\gamma$) to maintain compatibility with the visual-SLAM model. This also helps to reduce the computational overhead of maintaining a $6$D posecell network which is not required for navigation on a mobile robot platform.

\subsection{ViTa-SLAM}
In this section, we present the details of our novel visuo-tactile SLAM algorithm which we refer to as \textit{ViTa-SLAM}. 

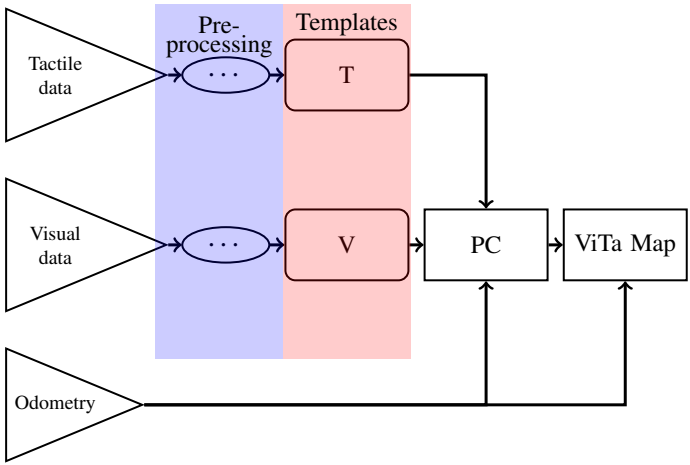
\begin{figure}[!htbp]
\centering
\resizebox{.5\textwidth}{!}{\input{sections/figures/vita_slam}}
\caption{Overview of the \textit{Vita-SLAM} architecture.}
\label{fig:vitaslam_architecture}
\end{figure}

The overall system architecture for \textit{ViTa-SLAM} is shown in Fig.~\ref{fig:vitaslam_architecture}: $3$ kinds of raw sensory data: tactile, visual, and odometry are now utilized simultaneously. Tactile and visual data are converted into visuo-tactile templates (T, V), respectively and hence, need to be pre-processed. A $3$D pose cell network is maintained. The experience in this approach is now defined as a $5$-tuple: $<x,y,\gamma,V,T>$ where $V$ is a visual template and $T$ is a tactile template at the 3D pose given by $x, y$ and $\gamma$. The experience map in this case will be referred to as \textit{vita map}. In contrast with the conventional experience map, the vita map's nodes contain visual and tactile data. The nodes are termed \textit{sparse} node if the tactile data is empty and \textit{dense} node otherwise. As an example, when the whiskers do not make contact, the whisker tactile information is not providing any information while the camera can still acquire novel scene information. When the whiskers are whisking a wall/landmark, both the camera and whiskers yield features that allow the creation of informative dense nodes which greatly help visuo-tactile SLAM. The properties of a vita-map node (dense or sparse) are stored in the vita-map but not further used.

\input{sections/vita-slam}

Algorithm \ref{alg:vita-slam} describes \textit{ViTa-SLAM} in more detail.
The visual and tactile processes are running continuously in parallel to the \textit{ViTa-SLAM} node, as shown by the $\bm\circlearrowleft$-symbol and store the current visual and tactile templates $V_{cur}$ and $T_{cur}$. 
The \textbf{visual process} follows the same steps as shown in Fig.~\ref{fig:view_template}. The resulting visual template is called $V_{cur}$.

Similarly, the \textbf{Tactile process} is pre-processing the input data consisting of the xy-deflection angles and the whisker contact points in world frame.
In line \ref{line:PFH}, the point feature histogram ($PFH$) is generated by creating a $N$-dimensional histogram of the contact points with $B$ bins per dimension.
The resulting histogram is then flattened into a $N^B$ histogram.
For each whisker, the slope of the xy-deflection between the initial contact to the maximum contact during one whisk cycle\footnote{One whisk cycle is defined as completing one full protraction/retraction cycle.} is computed. The result is the slope distribution array ($SDA$).
The current tactile template containing $PFH$ and $SDA$ is saved as $T_{cur}$. The two components can be extracted as $[t]_{PFH}$ and $[t]_{SDA}$.

In the \textbf{ViTa-SLAM} process, if a whisk cycle has been completed, the data stored in the visual and tactile processes, $V_{cur}$ and $T_{cur}$, is extracted. In line \ref{line:compare}, $V_{cur}$ and $T_{cur}$ are matched against all old visual and tactile templates and the id with the closest match $m$ and the corresponding error $\epsilon$ are returned. Finally, the error $\epsilon$ is used to determine if a novel template has been detected or a match with an old template has occurred. In either case, energy is injected at the template with match id $m$. After this process, the current visual and tactile templates are appended to the memory and the template match ID is published for experience map generation. 

The \textbf{template matching} function computes $\epsilon$ by comparing the current template to all visual and tactile templates in the memory. The visual error $V_{err}$ is computed as the pairwise sum of absolute differences between $V_{cur}$ and all visual templates in the memory. For the tactile data similarly, the PFH and SDA are treated separately and the respective errors ($PFH_{err}$ and $SDA_{err}$) are computed. A weighted sum of all obtained error terms yields the error $\epsilon_{cur}$ between the current visuo-tactile templates and the ones in the memory as:
\begin{equation}
\begin{split}
\epsilon_{cur} &= \alpha~|V_{cur}-v|_{L_1} + \beta~|PFH_{cur}-[t]_{PFH}|_{L_1}\\ &+ \gamma~|SDA_{cur}-[t]_{SDA}|_{L_1}\\
\textrm{where,}\\
\alpha &= \frac{1}{\sigma_V}\\
\beta &= \frac{1}{\sigma_{PFH}}\\
\gamma &= \frac{1}{\sigma_{SDA}}\\
\end{split}
\label{eq:combinederror}
\end{equation}
In Eq.~\eqref{eq:combinederror}, $|\cdot|_{L_1}$ represents the L1 norm between the corresponding terms.
$\alpha, \beta$ and $\gamma$ are scaling factors for the respective errors which represent the standard deviations of the raw sensory data. Finally, the returned match ID ($m$), is the ID of the combined template with the lowest $\epsilon$.

%% file: sections/figures/vanilla_ratslam.tex
\begin{tikzpicture}[
				round boxes/.style={draw, rectangle,%
                thick,minimum height=1cm, rounded corners,
                minimum width=1cm, black, text=black,
                text width=15mm, anchor=center, align=center},
                boxes/.style={draw, rectangle,%
                thick,minimum height=1cm,
                minimum width=1cm, black, text=black,
                text width=15mm, anchor=center, align=center}, 
   				right iso/.style={isosceles triangle,scale=0.8,sharp corners, anchor=center, xshift=-4mm},
    			left iso/.style={right iso, rotate=180, xshift=-8mm},
    			txt/.style={text width=1.5cm,anchor=center},
    			ellip/.style={ellipse,scale=0.8},
    			empty/.style={draw=none}
    			]
  \matrix (mat) [matrix of nodes, nodes=boxes, column sep=0.1cm, row sep=0.5cm] 
  {
         	   & |[ellip]|{View Template} & |[round boxes]|{V} & Bio-inspired PC & \\ 
	|[right iso]|{Visual data} & & & & Semi-Metric Exp. Map\\
         	    |[right iso]|{Odom.} & & & & \\ 
  };  
\draw [very thick, black, ->](mat-2-1)|-(mat-1-2);
\draw [very thick, black, ->](mat-1-2)--(mat-1-3);
\draw [very thick, black, ->](mat-1-3)--(mat-1-4);
\draw [very thick, black, ->](mat-1-4)-|(mat-2-5);
\draw [very thick, black, ->](mat-3-1)-|(mat-1-4);
\draw [very thick, black, ->](mat-3-1)-|(mat-2-5);

\node at(-2.8cm, 2.2cm) [right,fill=blue,text opacity=1,opacity=.2, minimum width=1.8cm, minimum height=2.4cm, text height=-1.6cm] {Preprocessing};

\node at(-0.57cm, 2.2cm) [right,fill=red,text opacity=1,opacity=.2, minimum width=1.825cm, minimum height=2.4cm, text height=-1.6cm] {Templates};

\end{tikzpicture}

%% file: sections/figures/lv_flow.tex
\begin{tikzpicture}[boxes/.style={draw, rectangle,%
                thick,minimum height=1cm, rounded corners,
                minimum width=1cm, black, text=black,
                text width=15mm, align=center}]
  \matrix (mat) [matrix of nodes, nodes=boxes, column sep=0.5cm, row sep=0.5cm] 
  {
         &           &               \\ 
	Crop & Subsample & Greyscale & Compare \\
         &           &               \\ 
  };  
\draw [very thick, black, ->](mat-2-1)--(mat-2-2);
\draw [very thick, black, ->](mat-2-2.east)--(mat-2-3.west);
\draw [very thick, black, ->](mat-2-3.east)--(mat-2-4.west);
\end{tikzpicture}

%% file: sections/figures/6dof_ratslam.tex
\begin{tikzpicture}[
				round boxes/.style={draw, rectangle,%
                thick,minimum height=1cm, rounded corners,
                minimum width=1cm, black, text=black,
                text width=15mm, anchor=center, align=center},
                boxes/.style={draw, rectangle,%
                thick,minimum height=1cm,
                minimum width=1cm, black, text=black,
                text width=15mm, anchor=center, align=center}, 
   				right iso/.style={isosceles triangle,scale=0.8,sharp corners, anchor=center, xshift=-4mm},
    			left iso/.style={right iso, rotate=180, xshift=-8mm},
    			txt/.style={text width=1.5cm,anchor=center},
    			ellip/.style={ellipse,scale=0.8},
    			empty/.style={draw=none,scale=0.01}
    			]
  \matrix (mat) [matrix of nodes, nodes=boxes, column sep=0.1cm, row sep=0.3cm] 
  {
         	 & |[ellip]|{$3$D Cts.} & |[round boxes]|{PFH~\cite{rusu2008learning}} &       &      \\ 
	|[right iso]|{Tactile data} 	 & & & |[round boxes]|{$T$} & PC \\
         	 & |[ellip]|{Defl.} & |[round boxes]|{SDA~\cite{kim2007biomimetic}}  & |[empty]| & Semi-Metric Exp. Map\\ 
    |[right iso]|{Odometry} & &  &  & \\
  };  
\draw [very thick, black, ->](mat-2-1)|-(mat-1-2); 
\draw [very thick, black, ->](mat-2-1)|-(mat-3-2); 
\draw [very thick, black, ->](mat-1-2)--(mat-1-3); 
\draw [very thick, black, ->](mat-3-2)--(mat-3-3); 
\draw [very thick, black, ->](mat-1-3)|-(mat-2-4); 
\draw [very thick, black, ->](mat-3-3)|-(mat-2-4); 
\draw [very thick, black, ->](mat-2-4)--(mat-2-5); 

\draw [very thick, black, -](mat-4-1)-|(mat-3-4); 
\draw [very thick, black, ->](mat-3-4)--(mat-2-5); 

\draw [very thick, black, ->](mat-2-5)--(mat-3-5); 
\draw [very thick, black, ->](mat-4-1)-|(mat-3-5); 

\node at(-2.8cm, 1.3cm) [right,fill=blue,text opacity=1,opacity=.2, minimum width=1.8cm, minimum height=6cm, text height=-5cm] {Preprocessing};

\node at(-0.6cm, 1.3cm) [right,fill=red,text opacity=1,opacity=.2, minimum width=3.8cm, minimum height=6cm, text height=-5cm] {Templates};

\end{tikzpicture}

%% file: sections/figures/vita_slam.tex
\begin{tikzpicture}[
				round boxes/.style={draw, rectangle,%
                thick,minimum height=1cm, rounded corners,
                minimum width=1cm, black, text=black,
                text width=15mm, anchor=center, align=center},
                boxes/.style={draw, rectangle,%
                thick,minimum height=1cm,
                minimum width=1cm, black, text=black,
                text width=15mm, anchor=center, align=center}, 
   				right iso/.style={isosceles triangle,scale=0.8,sharp corners, anchor=center, xshift=-4mm},
    			left iso/.style={right iso, rotate=180, xshift=-8mm},
    			txt/.style={text width=1.5cm,anchor=center},
    			ellip/.style={ellipse,scale=0.5},
    			empty/.style={draw=none}
    			]
  \matrix (mat) [matrix of nodes, nodes=boxes, column sep=0.2cm, row sep=0.5cm] 
  {
                 &   &   &           &           \\ 
    |[right iso]|{Tactile data} & |[ellip]|{\Huge\ldots} & |[round boxes]|{T} &           &           \\
    |[right iso]|{Visual data} & |[ellip]|{\Huge\ldots} & |[round boxes]|{V} & PC & ViTa Map \\
    |[right iso]|{Odometry}     &   &   &           &           \\
  };  
  
\draw [very thick, black, ->](mat-2-1)--(mat-2-2);
\draw [very thick, black, ->](mat-2-2)--(mat-2-3);
\draw [very thick, black, ->](mat-2-3)-|(mat-3-4);

\draw [very thick, black, ->](mat-3-1)--(mat-3-2);
\draw [very thick, black, ->](mat-3-2)--(mat-3-3);
\draw [very thick, black, ->](mat-3-3)--(mat-3-4);
\draw [very thick, black, ->](mat-3-4)--(mat-3-5);

\draw [very thick, black, ->](mat-4-1)-|(mat-3-4);
\draw [very thick, black, ->](mat-4-1)-|(mat-3-5);


\node at(-2.7cm, .5cm) [right,fill=blue,text opacity=1,opacity=.2, minimum width=1.8cm, minimum height=5cm, text height=-4.2cm] {Pre-};
\node at(-2.7cm, 0.25cm) [right,text opacity=1,opacity=.2, minimum width=1.8cm, minimum height=5cm, text height=-4.1cm] {processing};

\node at(-0.9cm, .5cm) [right,fill=red,text opacity=1,opacity=.2, minimum width=1.8cm, minimum height=5cm, text height=-4.2cm] {Templates};

\end{tikzpicture}

%% file: sections/vita-slam.tex
\begin{algorithm}
\label{alg:vita-slam}
\begin{algorithmic}[1]
\State $V_{old} \gets []$; $T_{old} \gets []$ \Comment{Template Memory}
\caption{Pseudocode for ViTa-SLAM\\}\label{alg:vita-slam}
\Algphase{Visual Process $\bm{\circlearrowleft$}}
\end{algorithmic}
\hspace*{\algorithmicindent} \textbf{Require:} RGB camera image $img$ \\
\hspace*{\algorithmicindent} \textbf{Output:} $V_{cur}$
\begin{algorithmic}[1]
\Function{visual\_template}{$img$}
\State $img \gets$\texttt{crop\_image($img$)}
\State $img \gets$\texttt{subsample($img$)}
\State $img \gets$ \texttt{to\_greyscale($img$)}
\State $V_{cur} \gets$\texttt{normalize\_image($img$)}
\EndFunction

\Algphase{Tactile Process $\bm{\circlearrowleft$}}
\end{algorithmic}
\hspace*{\algorithmicindent} \textbf{Require:} $Defl.$, $Cts.$\\
\hspace*{\algorithmicindent} \textbf{Output:} $T_{cur}$
\begin{algorithmic}[1]
\Function{tactile\_template}{$Defl., Cts.$}
\State $PFH \gets$ \texttt{multidim\_histogram($Cts.$)} \label{line:PFH}
\For{each whisker $w$}
	\State $init\_ct \gets [Defl._w > 0][0]$
	\State $max\_ct$ $\gets$ $\max(Defl._w)$
	\State $SDA_w$ $\gets$ \texttt{slope($init\_ct$, $max\_ct$)}
	
\EndFor
\State $T_{cur} \gets PFH \cup SDA$
\EndFunction

\Algphase{ViTa-SLAM $\bm{\circlearrowleft$}}
\end{algorithmic}
\hspace*{\algorithmicindent} \textbf{Require:} Template match threshold $\tau$, $whisk$\\
\hspace*{\algorithmicindent} \textbf{Output:} $m$
\begin{algorithmic}[1]
\Procedure{ViTA-SLAM}{}
\State $V_{cur}, T_{cur} \gets $ \texttt{read\_data()}
\If{$whisk$}
	\State $m, \epsilon \gets $ \texttt{COMP($V_{cur}, V_{old}, T_{cur}, T_{old}$)}\label{line:compare}
	\If{$\epsilon \leq \tau$}
		\State \texttt{inject($match\_id$)}
	\Else
		\State $m \gets$ \texttt{create\_template()}
		\State \texttt{inject($match\_id$)}
	\EndIf
	\State $V_{old} \cup V_{cur}$ \Comment{Append to memory}
	\State $T_{old} \cup T_{cur}$
\EndIf
\State \texttt{publish($m$)} \Comment{Publish match ID}
\EndProcedure

\Algphase{Template Matching}
\end{algorithmic}
\hspace*{\algorithmicindent} \textbf{Require:} $V_{cur}, T_{cur}$, $V_{old}, T_{old}$\\
\hspace*{\algorithmicindent} \textbf{Output:} \textit{$m$},  $\epsilon$
\begin{algorithmic}[1]
\Procedure{comp}{$V_{cur}, V_{old}, T_{cur}, T_{old}$}
\State $\bm\epsilon \gets []$
\For{$\forall \{{v,t}\} \in \{V_{old}, T_{old}\}$}
	\State $V_{err} \gets$ \texttt{v\_diff($V_{cur},v$)}\label{line:vision_like_ratslam}	
	\State $PFH_{err}, SDA_{err} \gets$ \texttt{t\_diff($T_{cur},t$)}\label{line:tactile}
	\State $\alpha \gets \frac{1}{\sigma_V}; \beta \gets \frac{1}{\sigma_{PFH}}; \gamma \gets \frac{1}{\sigma_{SDA}}$
	\State $\epsilon_{cur} \gets$ \texttt{error($\alpha, V_{err}, \beta, PFH_{err}, \gamma, SDA_{err}$)} 
	\State $\bm{\epsilon}~\cup \epsilon_{cur}$ 
\EndFor
\State $\epsilon, m \gets \argmin(\bm{\epsilon})$
\State \textbf{return} $\epsilon, m$
\EndProcedure

\end{algorithmic}
\end{algorithm}

%% file: sections/experiments.tex
\section{Experimental Setup}
In this section, we describe the operational environment and the robot platform that were used for empirical validation of the proposed ViTa-SLAM algorithm.

\subsection{Robot Platform}
\begin{figure*}[!htbp]
\centering
\hspace*{-1.3cm}
\begin{subfigure}[t]{.32\textwidth}
  \centering
  \includegraphics[scale=0.07625]{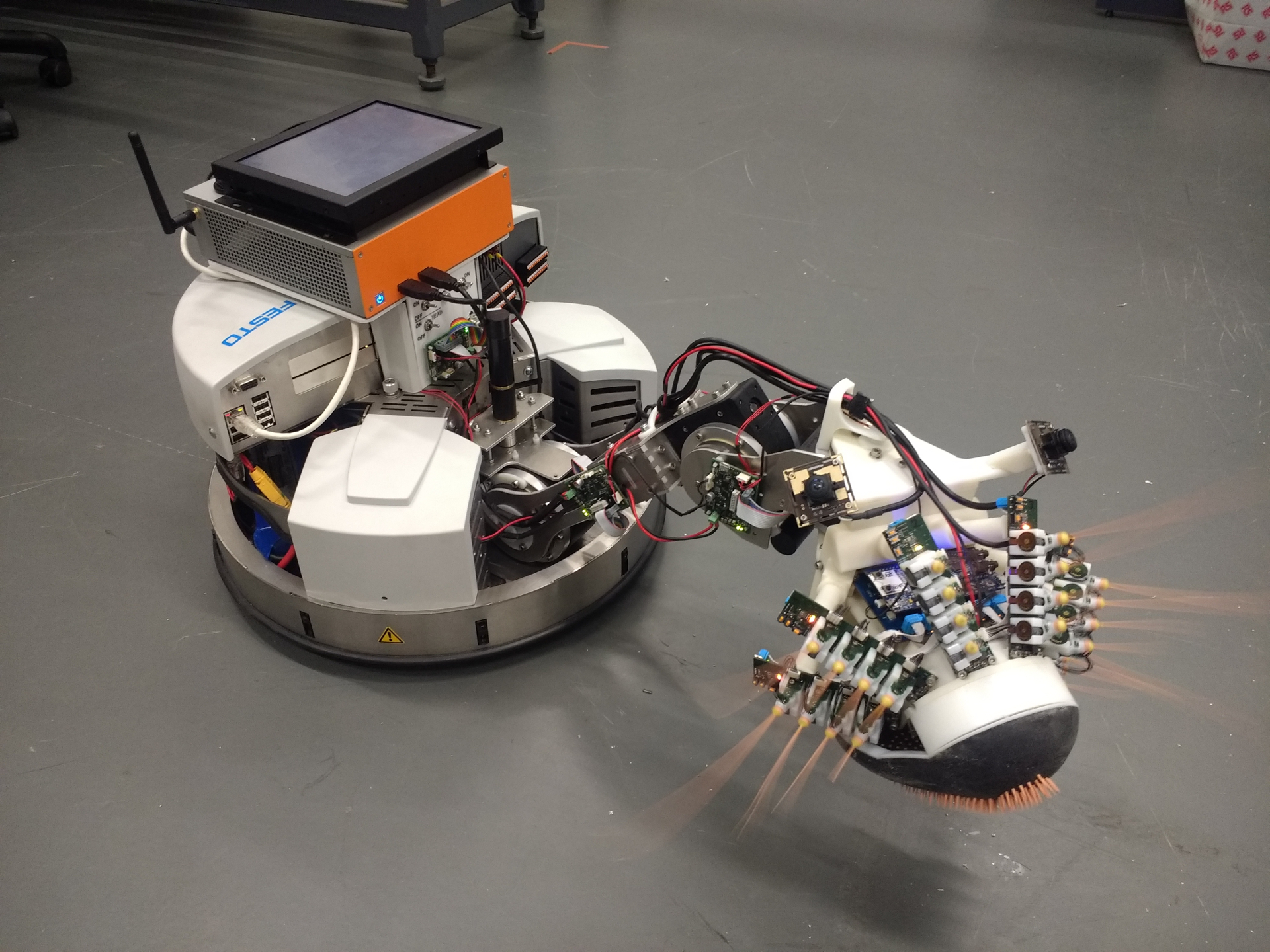}
        \caption{Physical platform.}
        \label{fig:real_whiskeye}
\end{subfigure}%
\hspace*{-0.4cm}
\begin{subfigure}[t]{.32\textwidth}
  \centering
  \includegraphics[scale=0.1275]{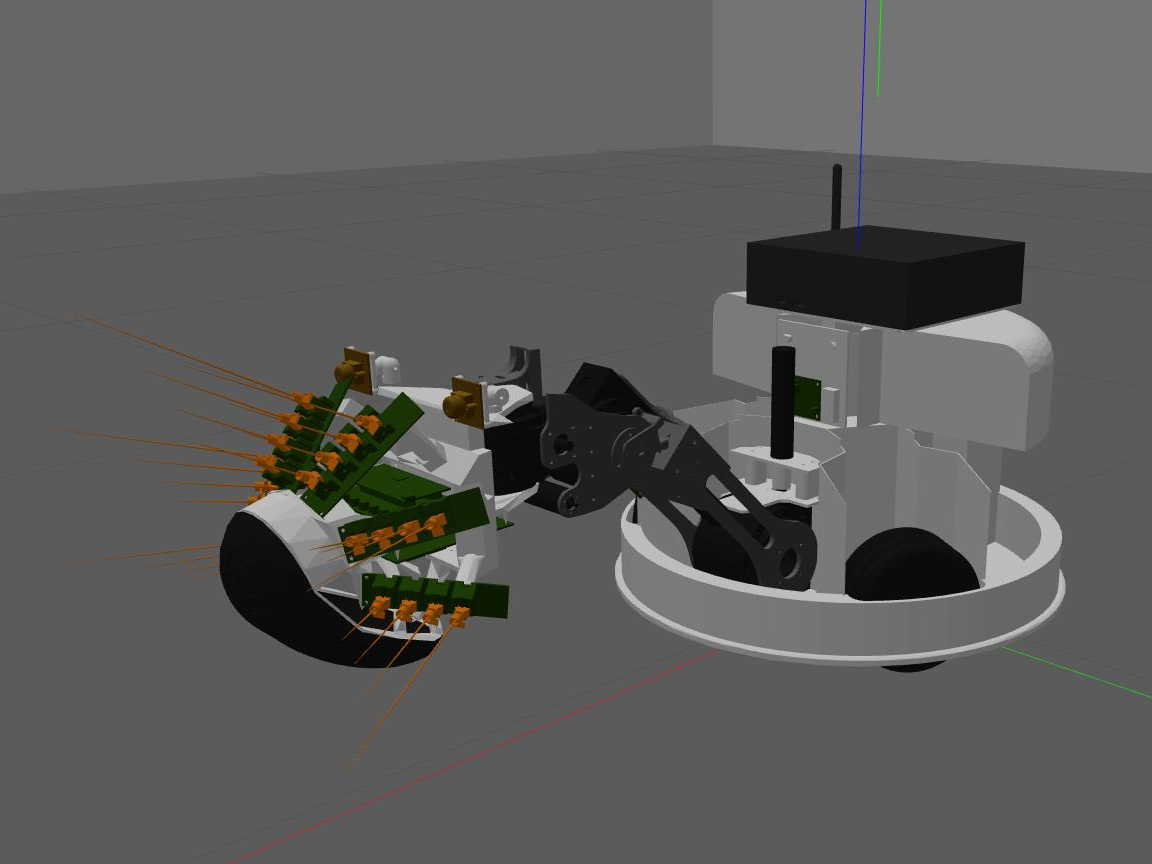}
  \caption{Simulated platform.}
  \label{fig:sim_whiskeye}
\end{subfigure}
\begin{subfigure}[t]{.32\textwidth}
	\centering
	\includegraphics[scale=0.084]{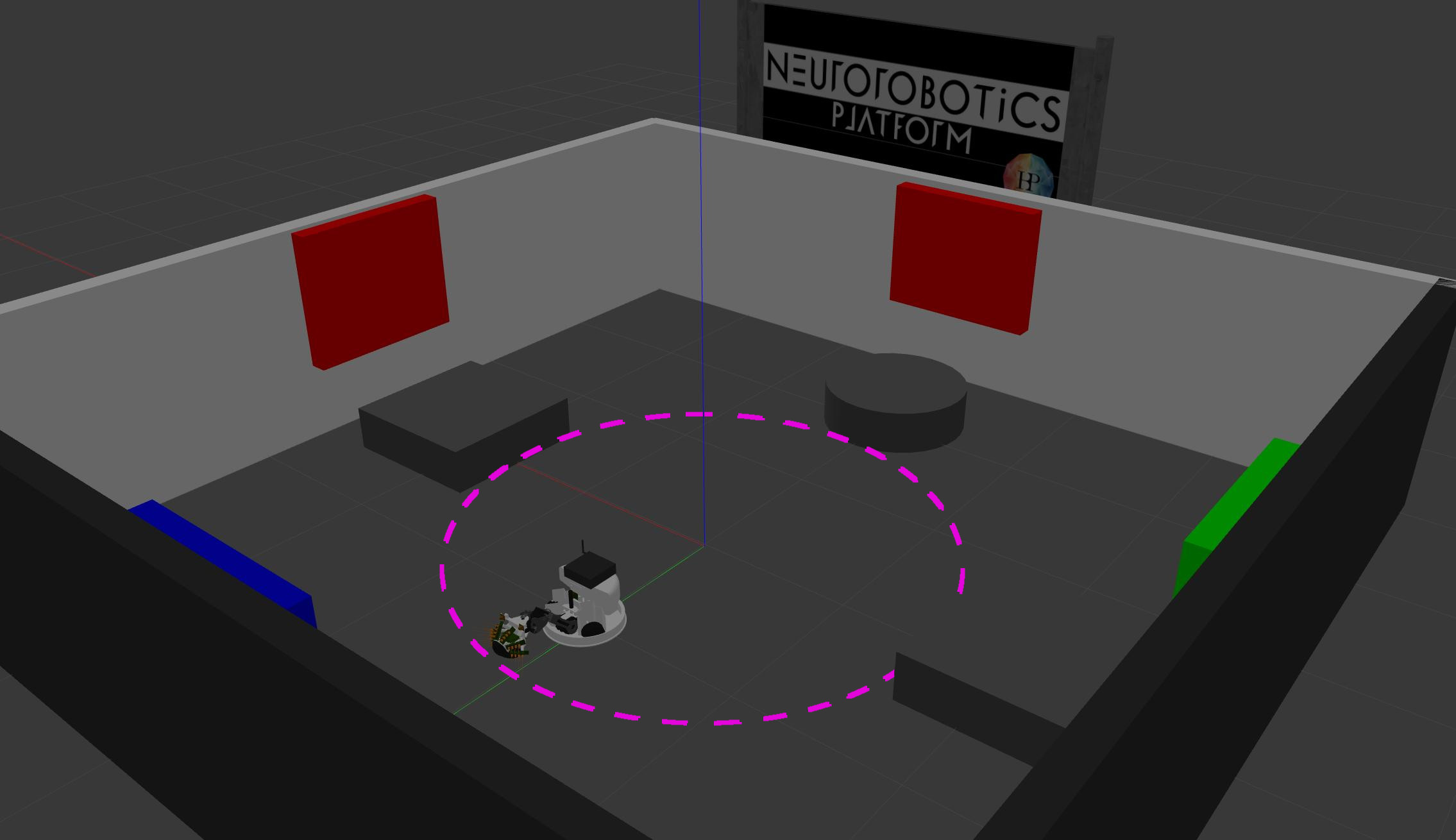}
	\caption{Operational environment.}
	\label{fig:operational_env}
\end{subfigure}
\caption{WhiskEye robot platform and its operational environment with the trajectory overlaid in magenta.}
\label{fig:robot_platform}
\end{figure*}

The robot platform used for this research is called the WhiskEye (Fig.~\ref{fig:real_whiskeye}) the design of which is based on a previous whiskered robot\cite{pearson2013simultaneous}. WhiskEye is composed of a Festo Robotino \textit{body}, a $3$ DoF \textit{neck}, and a $3$D printed \textit{head}. The robot is ROS compatible which allows for candidate control architectures to be developed and deployed on either the physical platform or the Gazebo simulation (shown in Fig.~\ref{fig:sim_whiskeye}) of WhiskEye as used in this study. Mounted on the head are the visual and tactile sensors. Two monocular cameras with a resolution of $640 \times 480$ pixels each provide a stream of RGB images with $5$ frames per second. An array of artificial whiskers consisting of $24$ \textit{macro-vibrissae} whiskers arranged into $4$ rows of $6$ whiskers provides tactile information. Each whisker is equipped with a $2$-axis hall effect sensor to detect $2$D deflections of the whisker shaft measured at its base during, and is actuated using a small BLDC motor to reproduce the active whisking behavior observed in rodents. The tactile data from whiskers is extracted during every whisk cycle, which takes $1$ second to complete.

\subsection{Operational Environment}
As a proof of concept, the algorithm was primarily tested in a simulated aliased environment to test visual-, tactile- and ViTa-SLAM under the challenging conditions a rodent faces in nature including: coarse vision, ill-lit tunnels, ambiguous visual and tactile environments. Fig.~\ref{fig:operational_env} shows the used simulated environment, a $6\times 6$ $m^2$ arena with $4$ wall-mounted visual and $3$ tactile landmarks designed to be qualitatively equivalent to the natural environment. In this setting, the robot was made to revolve around the center of the arena, with a radius of $1$ m whilst facing outwards to the walls.

%% file: sections/performance.tex
\section{Performance metrics}
The following metrics were used to evaluate the performance of the proposed ViTa-SLAM against the Visual-SLAM and Tactile-SLAM.

\subsubsection{Localization Error Metric (LEM)}
The localization error metric (LEM) measures the root mean squared error (RMSE) between the true pose and the estimated pose where the error is calculated separately for position and orientation. Thus,

\begin{align}
\begin{split}
LEM(\cdot) &= \sqrt{\frac{1}{n}\sum_{i=1}^{n}e_i^2} \,,\\
\textrm {where,} \\
e_i &= \hat{(\cdot)} - (\cdot) 
\end{split}
\label{eq:loc_rmse}
\end{align}

In Eq.~\eqref{eq:loc_rmse}, $\hat{(\cdot)}$ refers to estimated position (orientation) while $(\cdot)$ refers to true position (orientation).

\subsubsection{Experience Metric (ExM)}
The Experience metric (ExM) introduced in~\cite{salman2016advancing} provides a performance measure for algorithms like RatSLAM that produce semi-metric topological maps with loop closures. The ExM is comprised of the \textit{average rate of re-localization} (ARR) and the \textit{average rate of correct re-localization} (ARCR). The ARR is defined as the ratio of re-localizations to total number of experiences excluding the base set. The \textit{base set} is a set of initial experiences that has to be selected to define the main loop closure and to provide a reference for relocalization with future experiences. The ARCR is defined as the ratio of correct re-localizations to the total number of re-localizations. Higher values (close to $1$) for both factors indicate high certainty in the pose estimate. In order to determine if a re-localization is \textit{correct} or \textit{incorrect}, a threshold is used to compare the accuracy of the estimated pose to the ground truth pose. An experience following an incorrect re-localization is labeled as \textit{invalid} until a correct re-localization occurs. An experience following a correct re-localization is labeled as \textit{valid}.

\subsubsection{Energy Metric (EnM)}
In \cite{kummerle2009measuring}, energy metric was proposed as a generic metric for evaluation of a variety of SLAM algorithms. The SLAM performance was measured in terms of the energy required to transform the SLAM trajectory to the true trajectory. Let $N$ represent the number of relations between experiences in vita map and their corresponding sample points from the set of collected pose data. Then, $\delta_{i,j} = x_i \ominus x_j$ represents the transformation from node $x_i$ to $x_j$. If $T(\cdot)$ and $R(\cdot)$ represent the translation and rotation operations, then the energy metric (EnM) can be defined as:

\begin{equation}
EnM = \frac{1}{N} \mathlarger{\mathlarger{\sum}}_{i,j} T(\hat{\delta}_{i,j} \ominus \delta_{i,j})^2 +
R(\hat{\delta}_{i,j} \ominus \delta_{i,j})^2
\end{equation}

\section{Performance Evaluation}
\begin{figure*}[!htbp]
\centering
\begin{subfigure}{.32\textwidth}
  \centering
  \includegraphics[scale=0.38]{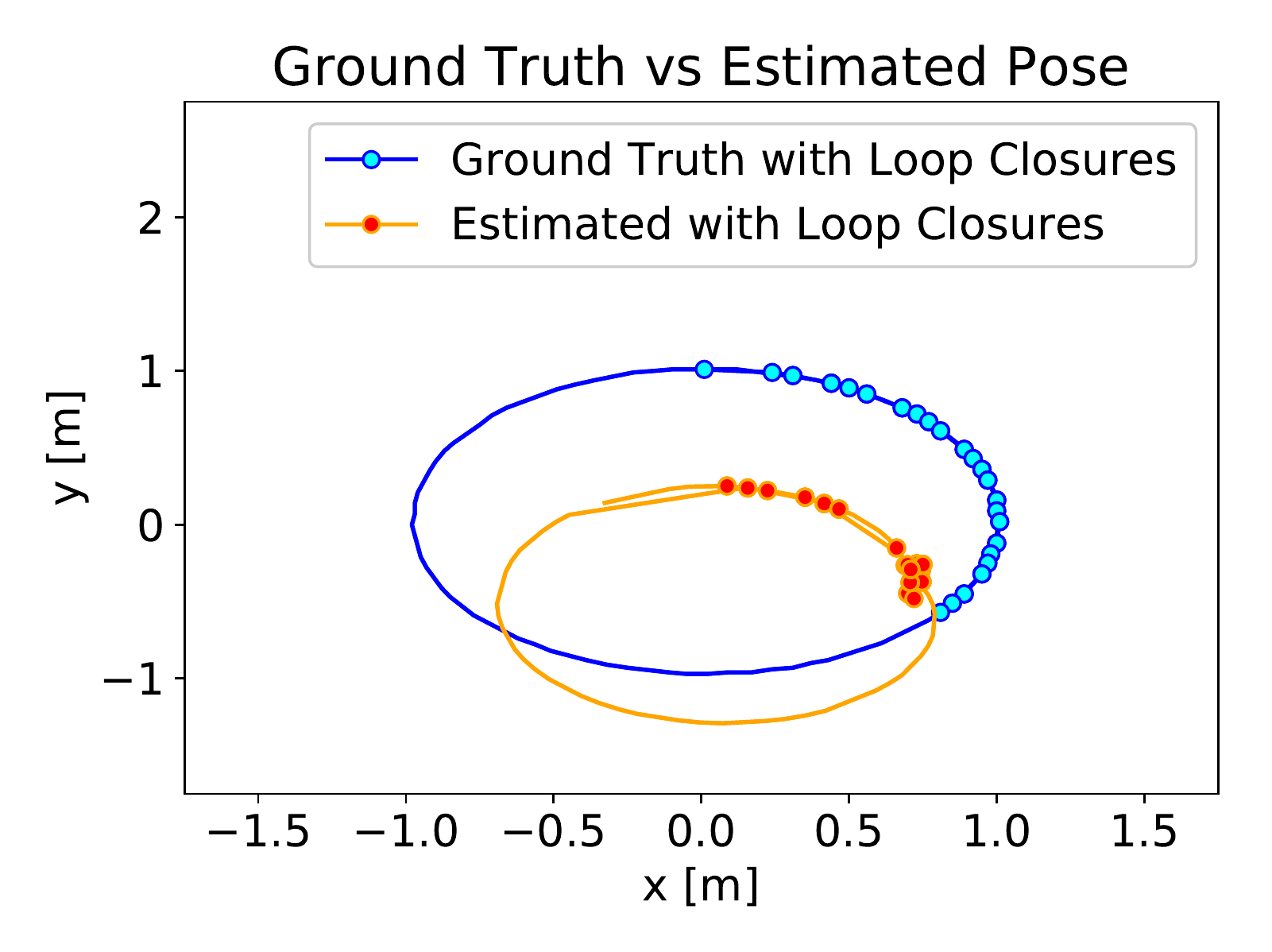}
  \caption{Visual-SLAM.}
\end{subfigure}%
\begin{subfigure}{.32\textwidth}
  \centering
  \includegraphics[scale=0.38]{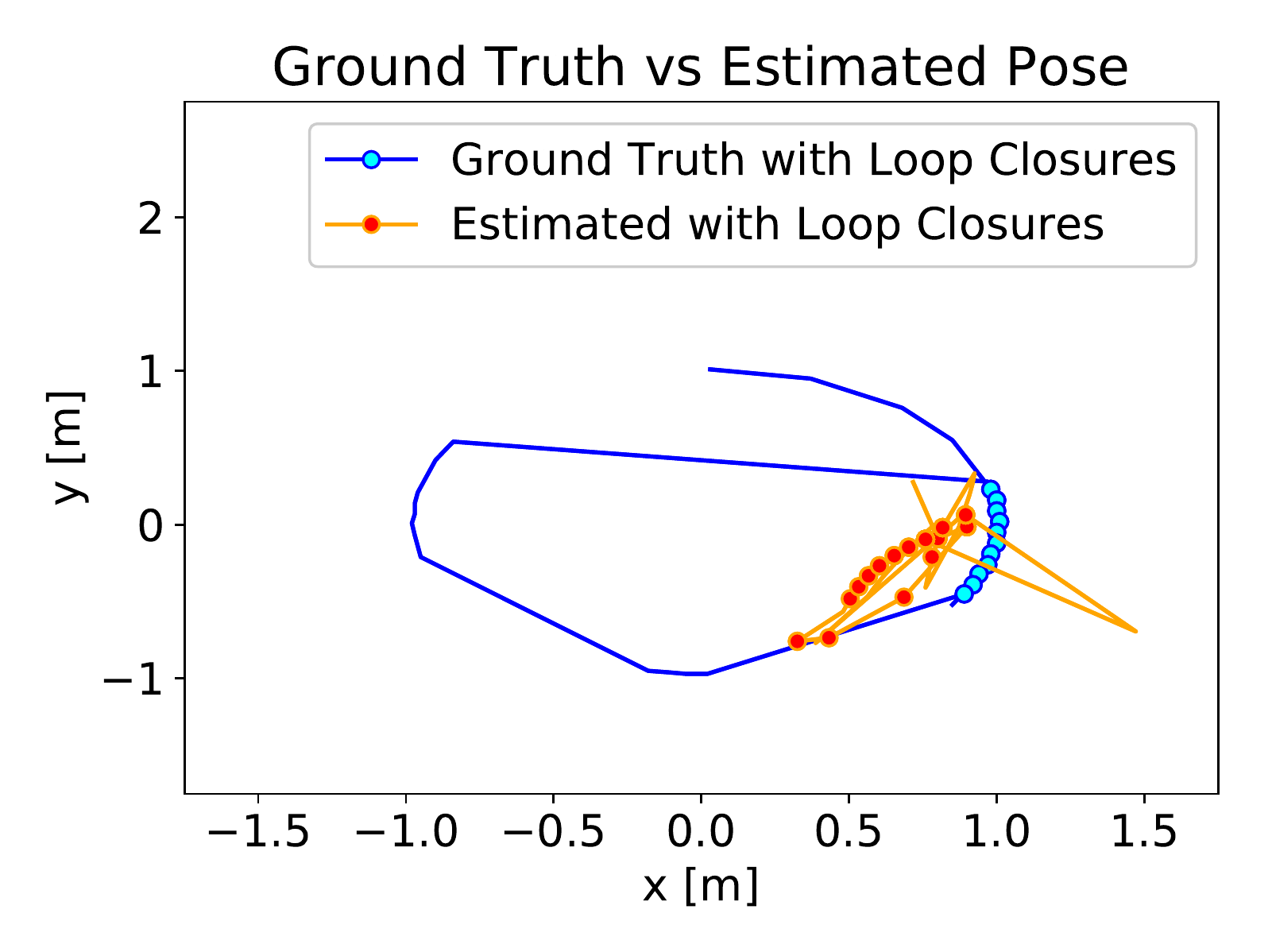}
  \caption{Tactile-SLAM.}
\end{subfigure}%
\begin{subfigure}{.32\textwidth}
  \centering
  \includegraphics[scale=0.38]{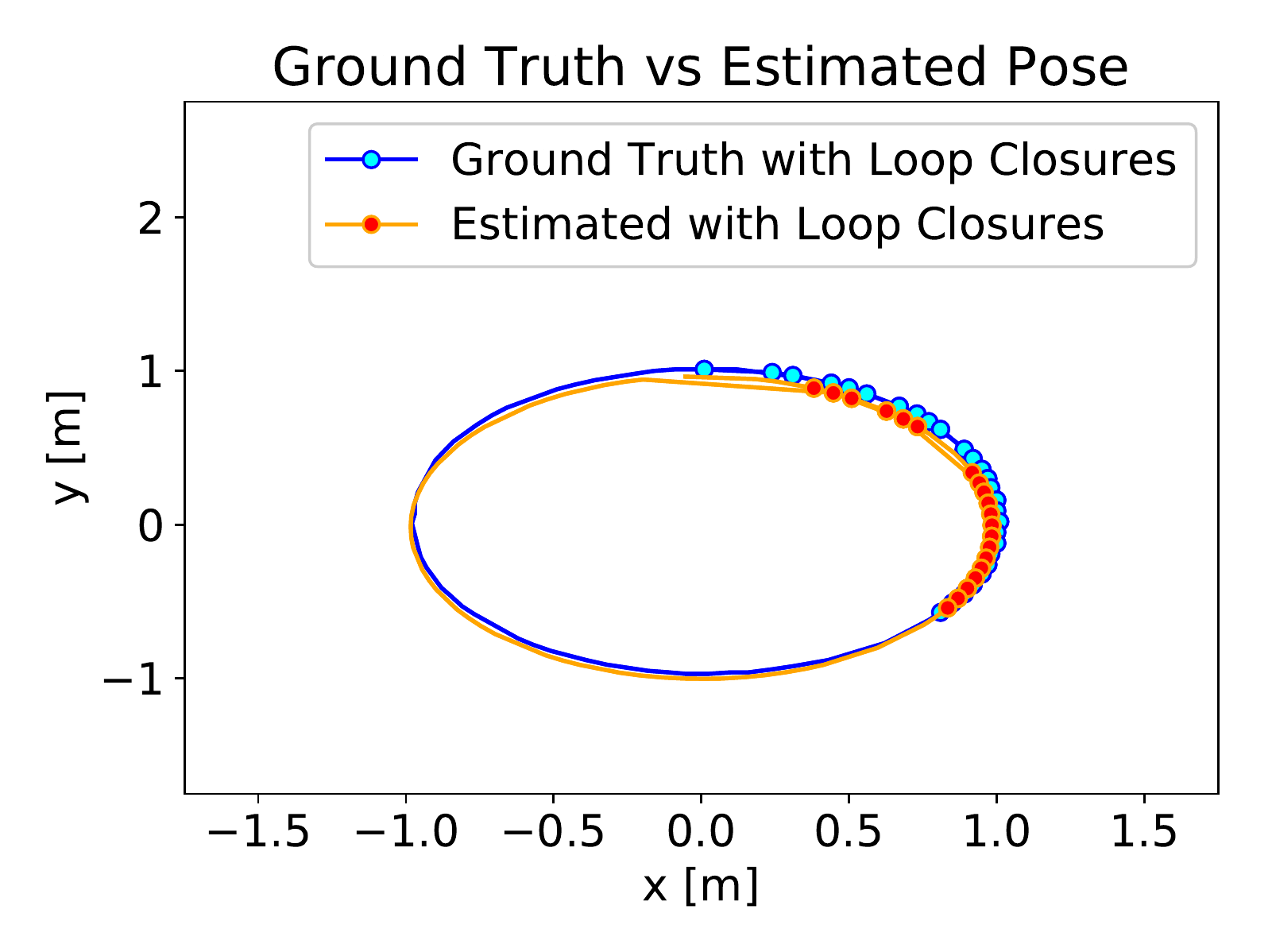}
  \caption{ViTa-SLAM.}
\end{subfigure}
\caption{Experience Map from various approaches.}
\label{fig:allExpMaps}
\end{figure*}

In this section, we compare the performance of \textit{ViTa-SLAM} against Visual and Tactile SLAM approaches using the metrics described above. The experience maps hence obtained are shown in Fig.~\ref{fig:allExpMaps} while the empirical summary of all the metrics is given in Table~\ref{tab:evaluation}.\footnote{Video demonstration available \href{https://www.youtube.com/watch?v=kiJn47GqczA&feature=youtu.be}{here}.}

The Energy Metric (EnM) shows that \textit{ViTa-SLAM} requires less energy to transform the trajectory to the ground truth by an order of magnitude. This confirms what can be seen from Fig.~\ref{fig:allExpMaps}, the experience maps of visual and tactile only SLAM are highly skewed as the result of wrong loop closures.

\begin{figure*}[!htbp]
\centering
\begin{subfigure}{.32\textwidth}
  \centering
  \includegraphics[scale=0.38]{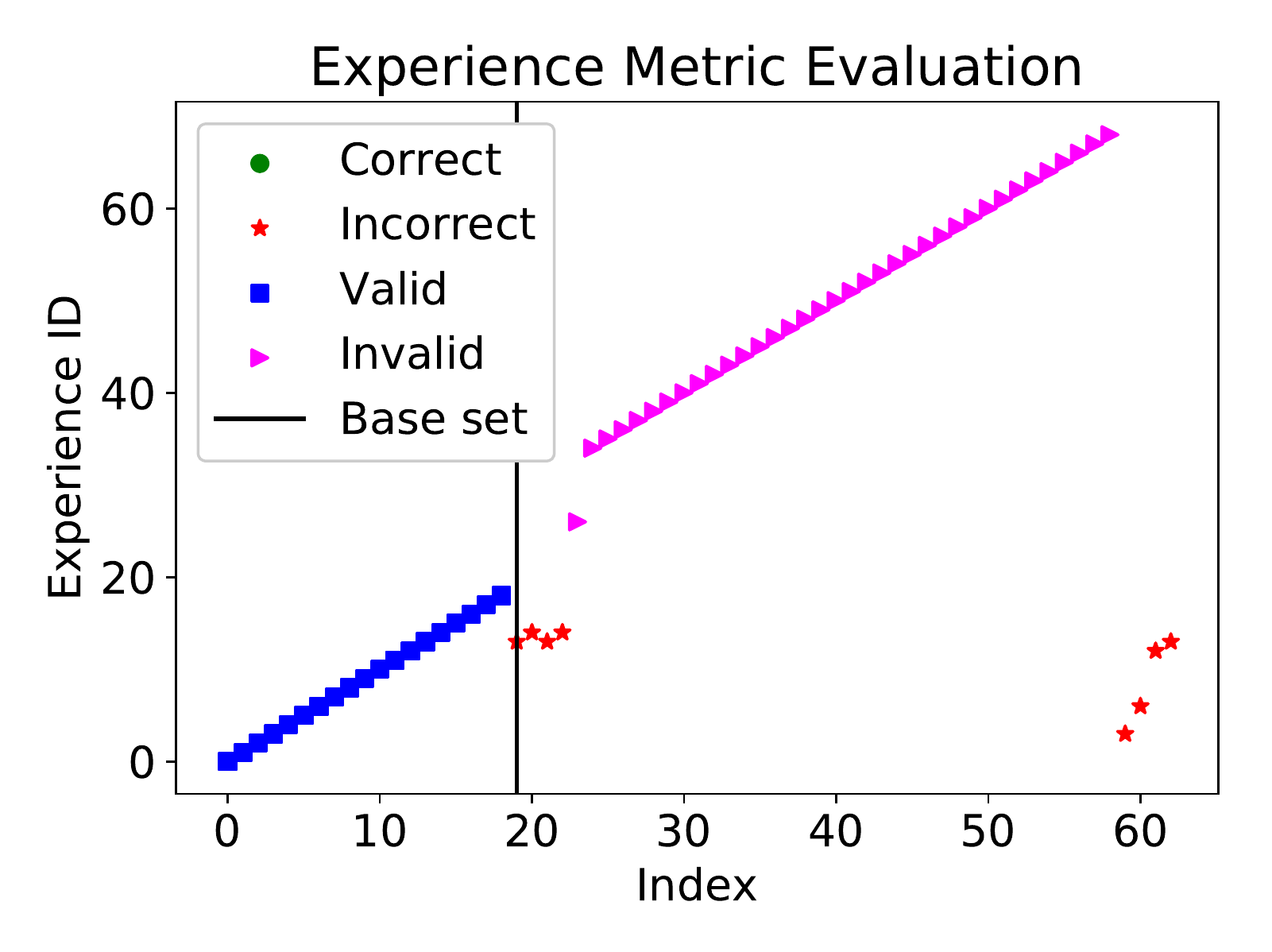}
  \caption{Visual-SLAM.}
  \label{fig:ExMa}
\end{subfigure}%
\begin{subfigure}{.32\textwidth}
  \centering
  \includegraphics[scale=0.38]{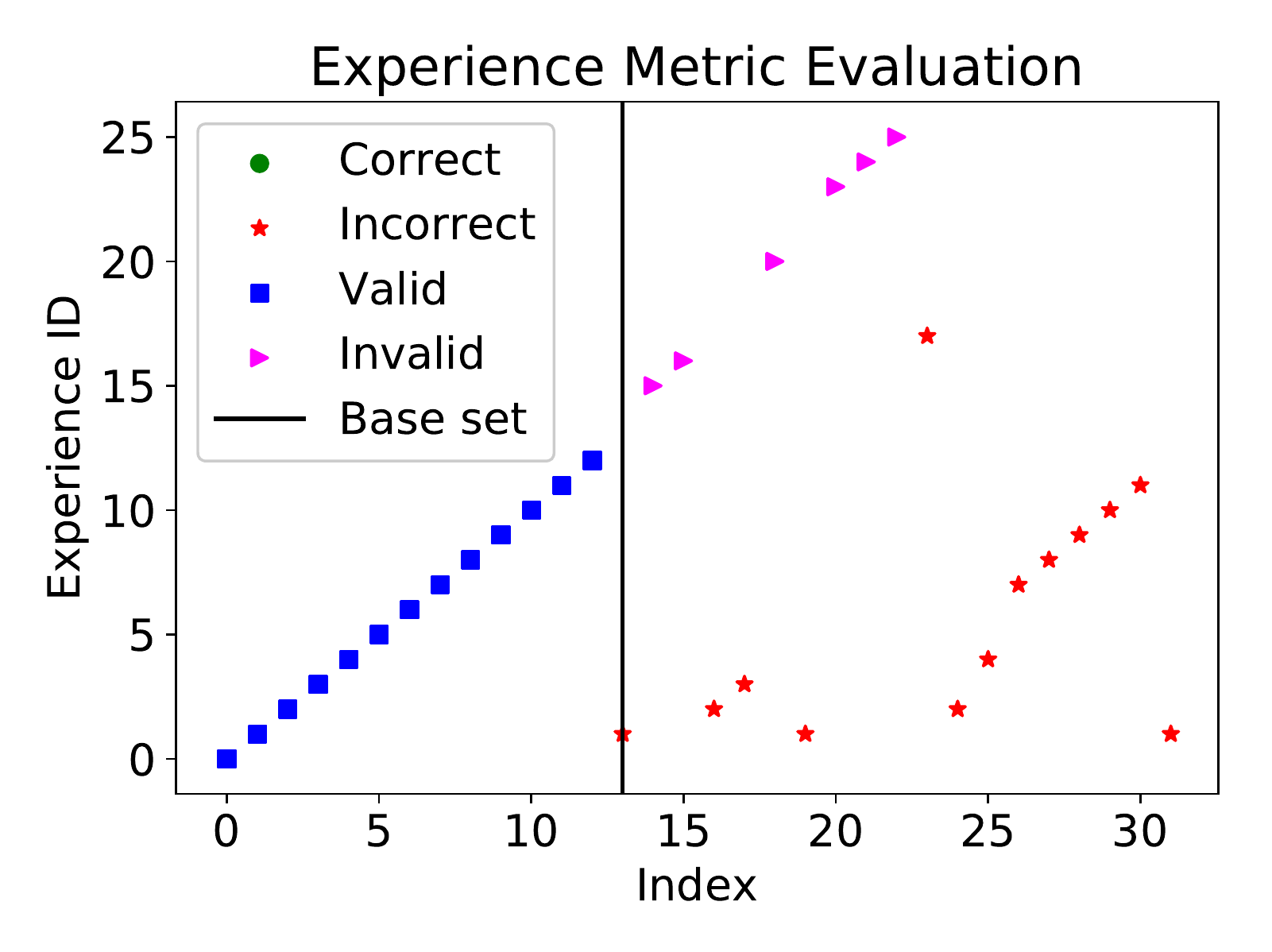}
  \caption{Tactile-SLAM.}
    \label{fig:ExMb}
\end{subfigure}%
\begin{subfigure}{.32\textwidth}
  \centering
  \includegraphics[scale=0.38]{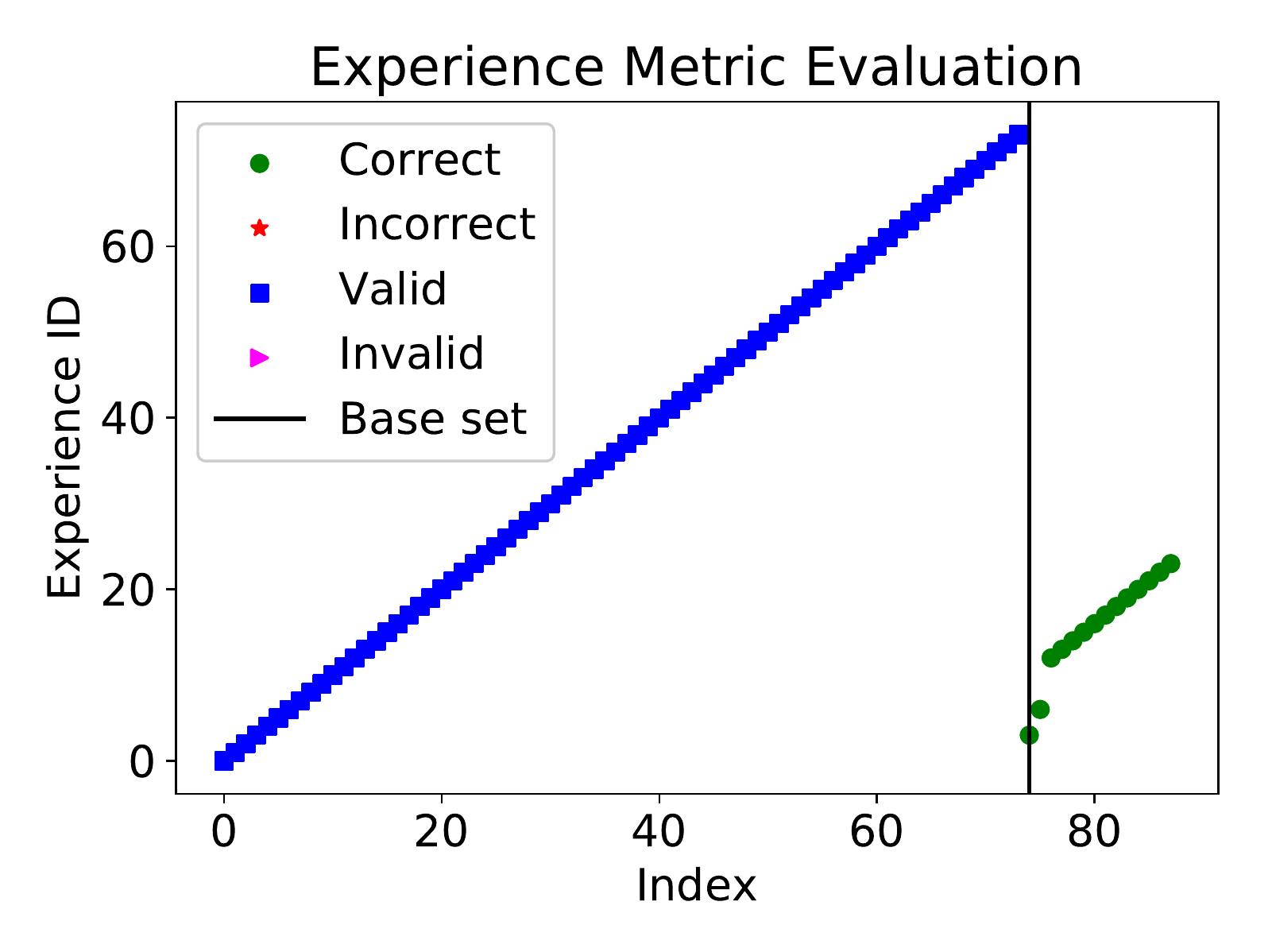}
  \caption{ViTa-SLAM.}
\end{subfigure}
\caption{Experience Metric (ExM) for all $3$ SLAM approaches.}
\label{fig:ExM}
\end{figure*}

To further evaluate the quality of loop closure detection we use the Experience Metric (ExM) with the thresholds for position and angular accuracy set to $0.08$ m and $4.6^\circ$ with the base set defined as the experiences generated during the first full rotation. The ARR and ARCR for \textit{ViTa-SLAM} show that \textit{ViTa-SLAM} is able to re-localize more often and correctly in all cases while the other methods always fail. The reason for this failure becomes evident in Fig.~\ref{fig:ExMa} and \ref{fig:ExMb}. For the visual- and tactile-SLAM methods, false positive re-localizations already occur in the base set as a result of the aliased environment. As opposed to this, \textit{ViTa-SLAM} successfully completes one rotation and correctly closes the loop.

The LEM further confirms these findings as the mean pose estimation accuracy of \textit{ViTa-SLAM} is clearly lower than the state-of-the-art methods.

\begin{table}[!htbp]
\caption{Performance Evaluation for RatSLAM, Whisker-RatSLAM, and ViTa-SLAM}
\label{tab:evaluation}
\centering
\resizebox{\columnwidth}{!}{%
\begin{tabular}{|c|c|c|c|c|c|}
\hline
 &\textbf{EnM} & \multicolumn{2}{c|}{\textbf{ExM}} & \multicolumn{2}{c|}{\textbf{LEM}} \\ \cline{2-6} 
\multirow{-2}{*}{\textbf{Method}} & \cellcolor[HTML]{000000} & \textbf{ARR} & \textbf{ARCR} &  \textbf{pos (m)} & \textbf{ori (rad)}\\ \hline
Visual-SLAM & $4.4024$  &$0.121$  &$0.0$  & $0.9168 \pm 0.6776$  &$1.8872 \pm 8.4155$\\ \hline
Tactile-SLAM  &$3.8129$  &$0.4643$  &$0.0$  & $1.1739 \pm 1.2901$  &$1.5604 \pm 3.0061$\\ \hline
ViTa-SLAM   &$0.4311$  &$0.7778$  &$1.0$ & $0.1445\pm0.0474$  &$0.6404 \pm 3.8371$\\ \hline

\end{tabular}
}
\end{table}

%% file: sections/conclusion.tex
\section{Conclusion and future works}
This work demonstrated a novel bio-inspired multi-sensory SLAM mechanism for a robot exploring and interacting with an environment that presents ambiguous cues. While previous attempts had been made to propose bio-inspired multi-sensory fusion, no prior research allowed for either environmental interactions through contact or fusion of unique and non-unique sensory information. To this end, ViTa-SLAM was presented which utilizes long-range visual and short-range whisker (tactile) sensory information for efficient bio-inspired SLAM. When comparing against earlier approaches that use only vision like RatSLAM or only tactile information like the Whisker-RatSLAM, it was shown that visuo-tactile sensor fusion can handle ambiguities that would otherwise lead to false positive loop-closure detection. However, similar to the previous methods, \textit{ViTa-SLAM} depends on hand-crafted features such as intensity profiles, point feature histogram and slope distribution array.

Therefore, we plan to improve the generalizability of \textit{ViTa-SLAM} by applying predictive coding to replace the hand-crafted features with learned features. We also plan to improve the robustness and acuity of the whisking behaviour whilst incorporating spatial attention mechanisms as is seen in rats \cite{mitchinson2013whisker}. Additionally, active spatial exploration strategies will be explored to improve the accuracy of localization and speed of mapping.